\documentclass[letterpaper, 10 pt, conference]{ieeeconf}  

\IEEEoverridecommandlockouts                              

\usepackage{graphicx}
\usepackage{amsmath} 
\usepackage{amssymb}  

\newcommand{\etal}{\textit{et al.}}
\usepackage{booktabs}

\usepackage{xcolor}

\usepackage[hidelinks]{hyperref}
\usepackage{comment}
\newcommand{\ra}[1]{\renewcommand{\arraystretch}{#1}}
\usepackage{multirow, makecell}

\title{\LARGE \bf
3D Scene Graph Prediction on Point Clouds Using Knowledge Graphs
}

\author{Yiding Qiu$^{1}$ and Henrik I. Christensen$^{2}$
\thanks{$^{1}$Yiding Qiu is with the Department of Computer Science and Engineering,
        University of California San Diego, La Jolla, 92122, USA
        {\tt\small yiqiu@eng.ucsd.edu}}%
\thanks{$^{2}$Henrik I. Christensen with the Department of Computer Science and Engineering,
        University of California San Diego, La Jolla, 92122, USA
        {\tt\small hichristensen@eng.ucsd.edu}}%
}

\begin{document}

\maketitle
\thispagestyle{empty}
\pagestyle{empty}

\begin{abstract}
3D scene graph prediction is a task that aims to concurrently predict object classes and their relationships within a 3D environment. As these environments are primarily designed by and for humans, incorporating commonsense knowledge regarding objects and their relationships can significantly constrain and enhance the prediction of the scene graph. In this paper, we investigate the application of commonsense knowledge graphs for 3D scene graph prediction on point clouds of indoor scenes. Through experiments conducted on a real-world indoor dataset, we demonstrate that integrating external commonsense knowledge via the message passing method leads to a $15.0\%$ improvement in scene graph prediction accuracy with external knowledge and $7.96\%$ with internal knowledge when compared to state-of-the-art algorithms. We also tested in the real world with 10 frames per second for scene graph generation to show the usage of the model in a more realistic robotics setting. 
\end{abstract}

\section{INTRODUCTION}

    

\begin{figure}[t]
   \centering
   \includegraphics[scale=0.58]{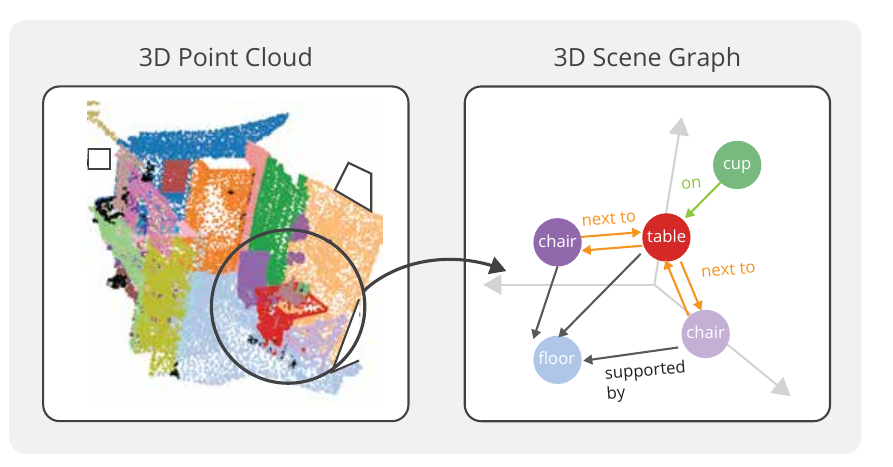}
   \caption{Given a 3D point cloud segmented into class-agnostic clusters, the objective is to generate the corresponding scene graph that labels the clusters and infers spatial or semantic relationships among them. The resulting 3D scene graph has nodes representing classes of objects, and directed labeled edges representing semantic relationships.}
   \label{fig:p1}
\end{figure}

A 3D scene graph is a high-level semantic scene representation that captures objects and their relationships within a 3D environment. This representation has recently demonstrated its potential for robotics tasks, including 3D scene reconstruction \cite{wu2021scenegraphfusion},  path planning \cite{agia2022taskography,amiri2022reasoning}, and navigation \cite{gomez2020hybrid,kim20193}. Most object-level semantic mapping methods primarily focus on predicting the class of objects in a scene~\cite{nuchter2008towards,sunderhauf2017meaningful}. However, 3D scene graph estimation diverges from these methods, as it requires additional tasks of (a) predicting if an edge should exist between two objects, and (b) predicting the label of the edge as a semantic relationship. Fig.~\ref{fig:p1} exemplifies the 3D scene graph prediction problem. Given point cloud data segmented into class-agnostic clusters, the task is to simultaneously classify the objects and relationships. The resulting scene graph comprises objects labeled as nodes and relationships labeled as directed edges. Multiple directed edges can exist from one object to another object. The graph can also be described using the triplet \textit{Subject-Predicate-Object}. To clarify the concept further, we follow the convention established in the 2D scene graph prediction domain, using the term "relationship" to describe the triplet and the term "predicate" to describe the label of the edge.

These relationships offer valuable information that benefits robotic tasks in multiple ways. First, semantic relationships can direct robots to search for target objects more efficiently. For instance, during object-goal navigation, the spatial relationship between two objects, such as \textit{Cup-On-Table}, can constrain the search space and improve object search efficiency. Secondly, labeled relationships offer robots a richer vocabulary to communicate with humans. For example, instead of describing a scene with the coordinates of a chair and a table, an agent can use the phrase \textit{Chair-NextTo-Table}, which resembles natural language more closely.

Compared to tasks focusing solely on object classification accuracy, scene graph prediction is more challenging, as accuracy is evaluated based on the correct prediction of triplets. One intuitive approach to address this problem is to utilize knowledge---the common structure or pattern typically found in most environments---to infer which object-relationship pairs are more likely to be present, given the current observation. Indoor environments have a human-centered structure and adhere to the laws of physics. For instance, chairs are usually situated near tables, and cups are commonly placed on tables or other large flat surfaces. Some triplets are more likely to exist than others, while some are nearly impossible to encounter in reality, such as \textit{Table-On-Cup}. Acquiring this knowledge from external resources or through training can simplify the prediction process. In some sense, knowing the predicates can help improve object recognition accuracy, and having higher confidence in object classes can increase or decrease the likelihood of some predicates.


We propose to incorporate external common-sense knowledge into 3D scene graph prediction tasks. The external knowledge graph is generated from various sources, including Visual Genome \cite{krishna2017visual}, ConceptNet \cite{liu2004conceptnet}, and WordNet \cite{miller1998wordnet}. Inspired by the Graph-bridging Network (GB-net) \cite{zareian2020bridging}, we use graph message passing methods to learn both node embeddings and edge embeddings within and between scene graphs and knowledge graphs and perform experiments on the indoor 3D scene graph dataset 3DSSG \cite{wald2020learning}. A major distinction between our work and \cite{zareian2020bridging} lies in the dataset type; \cite{zareian2020bridging} performs the task on 2D images for 2D scene graph prediction, whereas our task utilizes 3D point cloud data.  This presents a greater challenge because 3D segmented point clouds often suffer from missing points. In general, object recognition on point cloud data has lower accuracy as compared to images. Moreover, relationships in indoor environments are primarily spatial and geometric, while relationships in 2D image datasets can be conceptually abstract. Consequently, the types of knowledge graphs we employ differ from those used for 2D scene graph tasks. Additionally, indoor datasets typically have a smaller vocabulary for both objects and relationships, making it easier for common-sense knowledge to capture relationships.

To the best of our knowledge, our work is the first to leverage external common-sense knowledge for the 3D scene graph prediction problem. The current state-of-the-art algorithms are \cite{wald2020learning} and \cite{wu2021scenegraphfusion}.  However, neither of these methods utilizes external knowledge for prediction. We compare our prediction results with both works and demonstrate that our model outperforms both in scene graph prediction tasks. 

The main contributions of this paper are: (1) we proposed the model for 3D scene graph construction problem that incorporates common-sense knowledge and shows the improvement of the result. (2) we performed experiments in a real-world setting and demonstrate the possible application in the robotics domain. 

The remainder of the paper is organized as follows. In Section \ref{sect:relwrk}, we discuss related work, followed by the problem formulation in Section \ref{sect: task_def}. The main method and the network structure for our approach are described in Section \ref{sect:method}. In Section \ref{sect:exp_res} we discuss the dataset used, the overall experimental design, and the results. Finally, we summarize our work and outline future challenges in Section \ref{sec:conclusion}.




\section{Related Work}
\label{sect:relwrk}
\subsection{Scene graphs prediction with knowledge}
Scene graph prediction problems have been predominantly tackled on 2D image datasets. Two main methods are prevalent for integrating knowledge into the model. The first method is to extract the high-level structure inherent in the training dataset. For instance, MotifNet \cite{zellers2018neural} leverages the most frequent relations between labeled object pairs in the training set for scene graph prediction. Knowledge-embedded routing network (KERN) \cite{chen2019knowledge}, on the other hand, employs the co-occurrence probabilities between objects implicitly to aid in resolving the long-tail problem in the dataset. 

The second method involves utilizing external knowledge sources for the task. For example, Gu \etal \cite{gu2019scene} used ConceptNet \cite{liu2004conceptnet} to refine object and relationship features prior to training on scene graph generation. In \cite{zareian2020bridging}, Zareian \etal introduced GB-Net, which unifies scene graphs and knowledge graphs by learning the node encoding and edge encoding both within and between graphs. In their model, a scene graph represents an "image-conditioned instantiation" of a commonsense knowledge graph. They employed multiple knowledge bases as external knowledge sources, including WordNet \cite{miller1998wordnet}, ConceptNet \cite{liu2004conceptnet}, and Visual Genome \cite{krishna2017visual}. This represents one of the latest attempts to merge knowledge graphs and scene graphs, enabling learning of object features and predictors with neighboring nodes, while also constructing a "bridge" that infuses information between the two graphs through message passing.

Our work adopts the second method since we aim to generalize beyond the training data and deploy the algorithm in a real-world setting. we also aim to explore the effectiveness of external commonsense knowledge in generating 3D scene graphs, and as such, we adapt the bridging structure similar to \cite{zareian2020bridging}. However,  due to inherent differences in our dataset (image vs. point cloud), our method also takes into account the incompleteness of the point cloud dataset and innate 3D relationships in the reconstructed scene. Moreover, the knowledge graphs we construct differ from \cite{zareian2020bridging}., as the indoor robot task employs a distinct set of objects and relationships.

\subsection{3D scene understanding and scene graph generation}
The application of deep learning methods for 3D point cloud recognition can be traced back to PointNet \cite{qi2017pointnet}, a model still extensively to encode and predict point cloud data. In our study, we likewise use PointNet as a backbone for point cloud embedding.

Recently, there has been some research into indoor scene graphs specifically designed for indoor robot mapping. One example is the 3D scene graph \cite{armeni20193d},  which establishes a semi-automatic framework to create a dataset unifying objects, rooms, and cameras in a structured manner. While some traditional approaches \cite{rosinol20203d, hughes2022hydra} use SLAM for semantic mapping at the object level, their maps do not include labeled semantic relationships. Certain methods \cite{kim20193, amiri2022reasoning} construct a scene graph from each image, and then merge the 2D scene graphs into a global 3D graph. However, these approaches are tested on a small selection of objects and minimal relationships. The works most similar to ours include \cite{wald2020learning, wu2021scenegraphfusion}, with the 3DSSG dataset first introduced in these studies. \cite{wald2020learning} proposed the Scene Graph Spatial Network (SGPN), which learns the 3D scene graph from the reconstructed point clouds. \cite{wu2021scenegraphfusion} further extended the method to accommodate RGB-D images as input and demonstrated that a model initially trained on reconstructed point cloud could be used for online scene graph prediction with a minor decline in prediction performance.
\begin{figure*}[th]
\centering
   \includegraphics[scale=0.5]{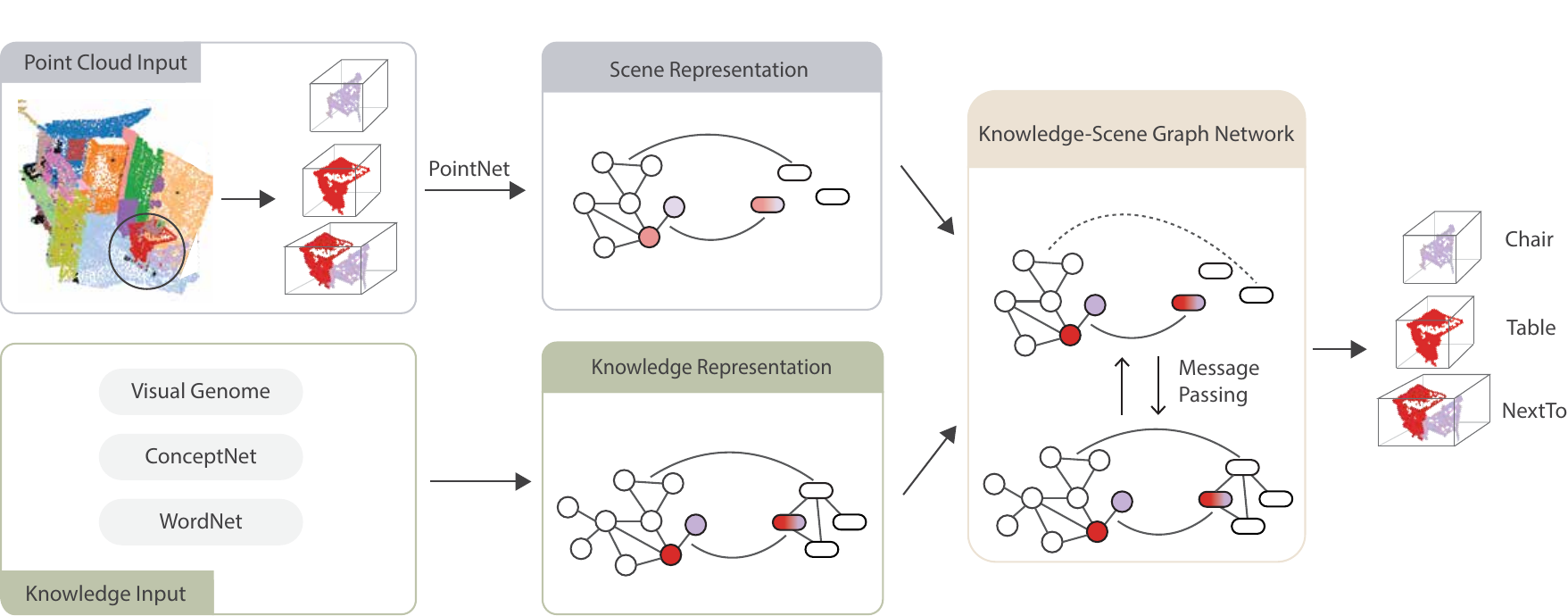}
   \caption{The pipeline of the overall model. The input is the class-agnostic point cloud of a scene. We begin by extracting point clouds such that each segment is either a subject or an object, and the predicate is the union of two segments. This forms the nodes of the Scene Representation(SR). Each node comprises a PointNet encoding feature and a contextual vector, with edges defined by distance. The knowledge input draws from three sources: ConceptNet, Visual Genome, and WordNet. The knowledge graph feature is the glove embedding, with edges constructed from these three sources. The scene graph and knowledge graph are subsequently trained together, allowing for simultaneous updates to both nodes and edges using the Knowledge-Scene Graph Network. Finally, the updated nodes from the intermediate scene graph are classified to establish relation triplets \textit{Subject-Predicate-Object}.}
   \label{fig:p2}
\end{figure*}

Based on these insights, our work focuses on generating scene graphs using the point cloud dataset, considering that the image-to-reconstructed point cloud can be accomplished either by traditional SLAM or by the graph fusing method. Notably, our approach differs from all previously mentioned methods by explicitly incorporating common-sense knowledge.


\section{Problem Formulation}
\label{sect: task_def}
Consider a constructed 3D scene with point cloud $\mathcal{P} \subset \mathbb{R}^3$ segmented into $n$ class-agnostic clusters $\mathcal{C}_i$ for $i = 1,\ldots,n$, and a directed graph $G = (\mathcal{V}, \mathcal{E})$. The objective of the task is to classify each cluster $C_i$ as an object $\mathcal{V}$ and associate predicates from $\mathcal{E}$ with it. The class of nodes and the directed edge eventually outputs a set of relation triplets \textit{Subject-Predicate-Object}. There are two forms to represent the 3D scene graph, which are semantically equivalent and can be transformed into each other. The triplet form is used for the evaluation of the model, whereas the graph form is utilized for map visualization and subsequent tasks, such as navigation.

\section{Method}
\label{sect:method}
\subsection{Pipeline}
As shown in Figure 2, the entire pipeline consists of two streams: one originating from the point cloud and the other from knowledge. The first stream generates the scene representation (SR) of the scene graph. Given the input of a room's point cloud, we extract the point cloud of each object and the predicate, in the form of the union of two clusters. These are encoded with a three-layer PointNet structure, constituting the first part of the SR node features. Additionally, we concatenate a set of contextual vectors to the PointNet feature embedding, as described in section \ref{sec:if}. Edges of the SR graph are kept if two objects in the 3d space are within a certain Euclidean distance threshold, with both directions preserved at this stage. 

The second stream originates from knowledge sources and thus is named knowledge representation(KR). Each node, representing either an object, subject, or predicate, is encoded using GloVe \cite{pennington2014glove} embedding. The edges are provided by three different knowledge sources: Visual Genome, Conceptnet, and WordNet. The process of constructing a knowledge graph is explained in detail in section \ref{sec:kg}. 

Next, both SR and KR are fed to a Knowledge-Scene Graph Network(KSGN). Through message passing, the node feature of both SR and KR will update alongside their neighboring node features, and the edge within and between graphs will update as well (Section \ref{sec:gbnet}). The resulting updated SR node feature of the object and predicate will finally pass through two layers of multi-layer perceptron(MLP) to predict relation triplets.

\subsection{Input feature}
\label{sec:if}
The nodes features of the input point cloud consist of the PointNet encoding and the contextual vector. The contextual vector is an 11-dimension vector first introduced in \cite{wu2021scenegraphfusion}. The vector includes the centroid of the point cloud $(x,y,z) \in \mathbb{R}^3$, the standard deviation that describes the sparsity of the segment $(\sigma_x,\sigma_y,\sigma_z) \in \mathbb{R}^3$, size of the bounding box $(b_x,b_y,b_z) \in \mathbb{R}^3$, maximum length of the segment $l = max(b_x,b_y,b_z) \in \mathbb{R}$, and bounding box volume $v = b_x \cdot b_y \cdot b_z \in \mathbb{R}$. The vector is calculated for each segment in the room as well as the union of two segments for the predicate. 
\label{tab:kg_table}
\begin{table}[hb!]
\renewcommand{\arraystretch}{1.7}
\begin{center}
\begin{tabular}{ccc}
Type            & Subtype   & Knowledge         \\ \hline
obj-obj         & subject-object       & VG         \\
(3 x 160 x 160) & object-subject       & VG         \\
                & relatedTo & ConcepNet  \\ \hline
obj-pred        & sub-pred  & VG         \\
(2 x 160 x 27)  & obj-pred  & VG        \\ \hline
pred-obj        & pred-sub  & VG         \\
(2 x 27 x 160)  & pred-obj  & VG        \\ \hline
pred-pred       & category  & hand label \\
(2 x 27 x 27)   & wup score & WordNet   
\end{tabular}
\end{center}
\caption{Knowledge graph adjacency matrix description}
\end{table}
\begin{figure*}[th!]
\begin{center}
   \includegraphics[scale=0.47]{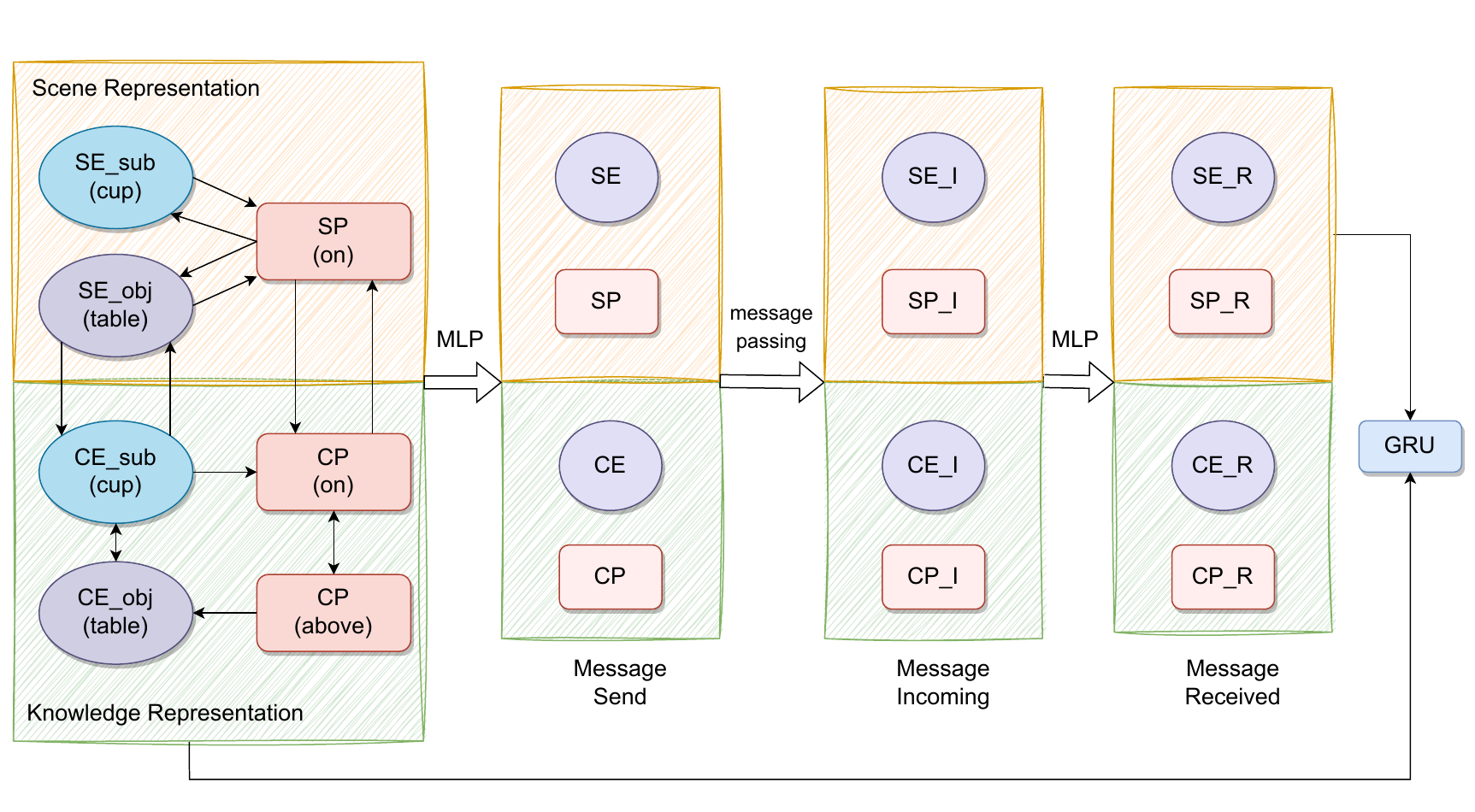}
   \caption{The pipeline of the graph bridging model}
   \label{fig:p3}
  \end{center}
\end{figure*}

\subsection{Knowledge Graph Construction}
\label{sec:kg}

The knowledge graphs are constructed from three different sources: 

\textbf{Visual Genome} \cite{krishna2017visual}: This contains labeled scene graphs for image datasets. We filtered the object and relationship vocabulary for the training data and constructed four different matrices: subject-object, object-subject, subject-predicate, and predicate-object. 

\textbf{ConceptNet} \cite{liu2004conceptnet}: A multilingual crowd-sourcing-based knowledge graph database. We retrieved an object-object matrix from it.

\textbf{WordNet}\cite{miller1998wordnet}: A lexical database that contains semantics explanations and synonyms for nouns, verbs, and adjectives. We use WordNet for predicate-predicate relationships. 

The knowledge graph in our study is defined using four types of adjacency matrices, as depicted in Figure \ref{fig:p2} and Table I. For each object in the dataset, we assume they can either be an object type or a subject type. Thus for both subject-object and object-subject relationships, the adjacency matrices are square matrices. We list the details of the knowledge graph in Table I. Because the dataset we use contains 160 objects and 27 relationships, we also listed the matrix dimension for clarity. 

Regrading ConceptNet, we use \textit{relatedTo} for between-object relationships, with a binary weight. As for the predicate-predicate relationship, we have created a manually crafted categorical adjacency matrix. This matrix connects all directional spatial relationships (such as left, right, behind, front), comparison relationships (smaller than, bigger than, higher than, lower than), and relationships that imply an attachment (attached to, standing on, lying on). This manually created matrix is binary as well. For determining similarity scores between predicates, we use WordNet to get the Wu-Palmer (WUP) scores. For all other adjacency matrices, weights are normalized in a range from 0 to 1.

\subsection{Knowledge-Scene Graph Network}
\label{sec:gbnet}
The Knowledge-based Scene Graph model (KSGN) is a modified version of Gb-Net \cite{zareian2020bridging}. As shown in Figure \ref{fig:p2}, KSGN accepts two types of graph input: the scene graph and the knowledge graph. Notably, both inputs are heterogeneous graphs.

The detailed model is depicted in Figure \ref{fig:p3}. The scene representation consists of Scene Entities(SE) and Scene Predicates (SP), and the common-sense knowledge representation consists of Common-Sense Entities (CE) and Common-Sense Predicates (CP). In the following, we use ${\Delta} \in (SE, SP, CE, CP)$ to represent different types of nodes. 

Each node is encoded by 2-layers of Multi-Layer Perceptron (MLP) to form the node features as the Message Send: 
\begin{equation}
    \mathbf{m}_i^{\Delta \rightarrow}=\phi_{\text {send }}^{\Delta}\left(\mathbf{x}_i^{\Delta}\right)
\end{equation}
where $\phi_{\text {send }}$ is the MLP that is named as "send head", It is trained and shares the weight across four types of nodes.

With each outgoing message, we compute the message along each incoming edge. This is done by first summing the weight of the same types of edges, and then concatenating across different types of edges.
\begin{equation}
\mathbf{m}_j^{\Delta \leftarrow}=\phi_{\text {receive }}^{\Delta}\left(\bigcup_{\Delta^{\prime}} \bigcup^{\mathcal{E}_k \in \mathcal{E}^{\Delta^{\prime^{\prime} \rightarrow \Delta}}}\sum_{\left(i, j, a_{i j}^k\right) \in \mathcal{E}_k}a_{i j}^k \mathbf{m}_i^{\Delta^{\prime} \rightarrow}\right)
\end{equation}
Finally, given the original input nodes and the received message, we use Gated Relu Unit (GRU) to update the node representations. The updated node vector is used to classify objects and predicts through training and back-propagation. 

\label{tab:table1}
\begin{table*}[tbh!]
  \begin{center}
  {\small
  \ra{1.2}{
\begin{tabular}{l cccc cccc}
\toprule
 & \multicolumn{4}{c}{$O_{160}R_{26}$} & \multicolumn{4}{c}{$O_{27}R_{7}$}\\
\cmidrule(lr){2-5} \cmidrule(lr){6-9}
Model & RE $\uparrow$ & RE$_{single}$ $\uparrow$ & Obj@1$\uparrow$ & Obj@5 $\uparrow$ & RE $\uparrow$ & RE$_{single}$ $\uparrow$ & Obj@1 $\uparrow$& Obj@5 $\uparrow$ \\ \hline
\midrule
SGPN \cite{wald2020learning}
&0.071
&0.119
&0.357
&0.623
&0.383
&0.385
&0.420
&0.780\\
SGFN\cite{wu2021scenegraphfusion} 
&0.113
&0.169
&\textbf{0.504}
&\textbf{0.754}
&0.417
&0.417
&0.624
&\textbf{0.923}\\
\textbf{Ours} (internal KG) 
&0.122
&0.184
&0.466
&0.739
&0.450
&0.450
&\textbf{0.644}
&0.917\\
\textbf{Ours} (external KG) 
&\textbf{0.130}
&\textbf{0.187}
&0.473
&0.742
&\textbf{0.469}
&\textbf{0.470}
&0.637
&0.922\\
\bottomrule
\end{tabular}}}
\end{center}
\caption{Quantitative results of the evaluated methods in the recall. }
\end{table*}

\subsection{Loss}
The model is trained in an end-to-end fashion, and the total loss consists of the classification loss for both objects and predicates:
$$\mathcal{L}_{\text {total }}=\lambda \mathcal{L}_{\text {obj}}+\mathcal{L}_{\text {pred}}$$
where $\lambda$ is a user-defined weight factor. Because a subject can have multiple relationships with an object, we used cross-entropy loss for the $\mathcal{L}_{\text {pred}}$.


\section{Experiment on dataset}
\label{sect:exp_res}
\subsection{Task description}
The task objective is twofold: to generate scene graphs by (1) predicting the labels of segmented clusters of point cloud, and (2) predicting the labels of relationships between two point cloud clusters. For this, we use the 3RScan\cite{wald2019rio} dataset, which consists of 1482 scans across 478 different scenes. Each scene is recorded using Google Tango, yielding sequences of RGB-D images with accurately calibrated camera poses. The ground truth scene graph annotations are provided by 3DSSG \cite{wald2020learning}.

The entire dataset is divided based on the number of scans, including 1061 scenes for training, 117 for validation, and 157 for testing. The original dataset contains 534 object classes and 40 relationship classes. The type of relationships captured in the data include supporting relationships (e.g. standing, lying), proximity relationships (e.g. next to, in front of), and comparative relationships (e.g. bigger than, taller than). In this project, we trained and evaluated the algorithm in two different settings. The first setting consists of 160 object classes and 26 predicate classes ($O_{160}R_{26}$), and the second setting consists of 27 object classes and 7 predicate classes($O_{27}R_{7}$).

\subsection{Comparison Models}
\textbf{Scene Graph Spatial Network} (SGPN) \cite{wald2020learning} The network takes as input the object point cloud and the predicate point clouds and uses Graph Neural Network(GNN) to predict the scene graph.

\textbf{Scene Graph Fusion Network} (SGFN) \cite{wu2021scenegraphfusion} The network uses only contextual vectors as predicate input instead of PointNet encoded feature. It uses Graph Attention Network(GAT) among the triplets for the prediction. 

\textbf{Ours} We present two versions of models, one uses internal knowledge graphs, and the other use external knowledge graphs. Both models have the same structure. The only difference is that for the internal knowledge graphs, we initialize the knowledge representation with zero matrices, which allows the model to capture the relationship within the dataset. 

\subsection{Implementation details}
The model is implemented in PyTorch. The learning rate is set to 0.001. The input graphs are generated based on the distance between entities (0.5 meters). We use 100K epochs for our models and 300K epochs for SGPN and SGFN. The weight loss factor $\lambda$ is 0.5.

\subsection{Evaluation metric}
To evaluate the model's performance on scene graph generation, we compare the predicted \textit{Subject-Predicate-Object} triplets with the ground truth triplet. Considering that multiple predicates can co-exist between two objects, we employ two measures for evaluation: \textbf{RE} (abbreviation for relationship): This measure evaluates if the predicted triplets exactly match the ground truth. \textbf{RE$_{single}$}: This is a relaxed form of evaluation where at least one predicted relationship matches the ground truth.

Most of the current 2D scene graph prediction problems adopt recall rate as the evaluation metric \cite{lu2016visual, zellers2018neural, gu2019scene, zareian2020bridging}. Note that $Recall = TP/(TP+FN)$, where $TP$ is true positive and $FN$ is false negative. The major reason is that the ground-truth annotations of relationships are likely to be incomplete, and thus using metrics like accuracy or precision, which penalize false positive predictions, is unfair to reflect the actual performance of a model. In the table, we use Obj@K to represent if the prediction for objects exists in top K predictions.  

\begin{figure}[th!]
\begin{center}
   \includegraphics[scale=0.7]{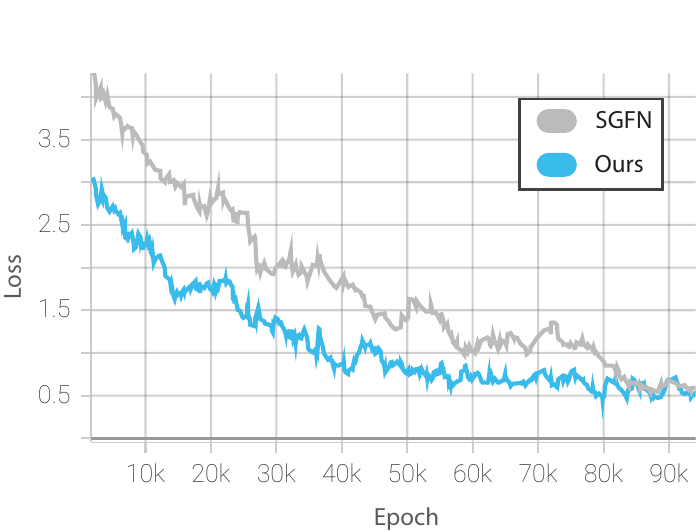}
   \caption{Convergence rate of models}
   \label{fig:p4}
  \end{center}
\end{figure}
\subsection{Results}
Table II summarizes the qualitative results for our model in comparison with state-of-art models. Our model outperforms in triplet prediction. The results indicate that incorporating knowledge graphs enhances overall performance, with external knowledge graphs offering greater improvements than internal knowledge graphs. Nevertheless, in terms of object classification accuracy, our model does not always give better predictions. This may be attributed to the inherent challenge of classifying objects using point cloud data, which in turn complicates the association between scene graphs and knowledge graphs for objects.

Fig \ref{fig:p4} illustrates how the use of knowledge graphs can expedite model convergence. When compared to the SGFN model, our model reaches convergence in earlier epochs, implying a reduced need for training epochs.

\subsection{Error Analysis}
We performed error analysis for both object classification and predicate classification. The top five misclassifications for objects, denoted as \textit{ground truth/prediction}, were \textit{wall/curtain}, \textit{wall/wardrobe},
\textit{wall/blinds},
\textit{pillow/cushion} and \textit{chair/side table}. The difficulty in recognizing curtains and blinds was also a common challenge faced during semantic classification for the ScanNet point cloud dataset \cite{dai2017scannet}. Regarding predicate classes, the most frequently occurring mistakes were \textit{learning against/close by}, \textit{cover/lying on}, and \textit{cover/standing on}.

\section{Real-world Experiment}
We conducted real-world experiments to assess both the limitations of our algorithm and its potential applications for robotics. While the 3RScan dataset is based on real-world data and employs RGB-D and inertial measurement unit(IMU) data from a Google Tango cell phone, there are hurdles to directly utilizing the trained model in real-time scenario. These challenges are primarily due to the requirement of offline post-processing for constructing the segmented 3D point cloud from RGB-D and IMU inputs, and the reliance on manually cleaned segmentation. Moreover, the camera poses provided by the original data was generated offline, providing a higher degree of accuracy than online camera pose estimates.

For our real-world experiments, we used the Intel RealSense D435 and RealSense T265 sensors. The camera poses and trajectories were estimated online using RTAB-Map\cite{labbe2019rtab}. Similar to \cite{wu2021scenegraphfusion}, we adopted the online segment fusion method from Tateno \etal \cite{tateno2015real}. This algorithm performs image segmentation on depth images and fuses point clouds given camera poses. In addition, while the model was trained on GPU, we employed Open Neural Network Exchange (ONNX) \cite{bai2019} in our real-world experiments to enhance inference speed and eliminate the need for a GPU. We conducted our scene graph generation experiments in two locations—an office area and a basement containing a kitchen—within a school building. Using RGB-D images, camera poses, and online image segmentation, the algorithm predicted the class of each point cloud and the relationships between segments for each frame, subsequently fusing this information on-the-fly. The result was a constructed 3D point cloud with annotated scene graphs on segments.

The results indicate that our model is particularly adept at detecting larger structures and furniture, such as walls, floors, cabinets, tables, and chairs. However, due to the sparsity of point clouds associated with smaller objects, segmenting and classifying these objects remains challenging. Our model operated in real-time, achieving more than 10 frames per second on average. The most frequently predicted accurate triplets are \textit{Wall-AttachedTo-Floor} and \textit{Chair-AttachedTo-Floor}. 

Potential improvements include the utilization of a more refined image segmentation algorithm, which could result in better 3D point cloud segmentation, albeit with a potential trade-off in inference speed. Additionally, the depth images provided by RealSense are somewhat noisy, suggesting that implementing filtering algorithms could help smooth the depth data, thereby improving segmentation.


\section{Conclusion}
\label{sec:conclusion}

In this study, we used external knowledge sources from Visual Genome, Conceptnet, and WordNet to predict the 3D scene graph on point cloud data constructed from RGB-D images. Our findings reveal that the use of external knowledge enhances the accuracy of scene graph prediction and expedites model convergence. We also conduct real-world experiments with the algorithm, demonstrating its capability to generate scene graphs online in a cluttered environment.

Nonetheless, our method exhibits several limitations. The primary challenge in 3D scene graph prediction lies in the accuracy of object classification on point clouds. One potential avenue for improving object recognition could involve the use of RGB-D images projected onto point clouds. Another limitation pertains to our current lack of use of common-sense knowledge regarding spatial and size relationships between objects. Given that point cloud data can be quite sparse for smaller objects, we could potentially leverage larger-sized objects and room information to improve the classification accuracy for smaller objects. As a future research direction, we plan to investigate various types of common-sense knowledge that can be used for scene graph prediction.











\clearpage

\bibliographystyle{IEEEtran}
\bibliography{reference}

\end{document}